\pdfoutput=1
\documentclass[11pt]{article}

\usepackage{naacl2021}
\usepackage{times}
\usepackage{latexsym}

\usepackage[T1]{fontenc}
\usepackage[utf8]{inputenc}

\usepackage{microtype}

\usepackage{booktabs}
\usepackage{makecell}
\usepackage{tabularx}
\usepackage{tabu}% http://ctan.org/pkg/tabu
\usepackage{xcolor}% http://ctan.org/pkg/xcolor

\usepackage{amsmath}
\usepackage{amssymb}
\usepackage{bm}
\newcommand{\bs}{\boldsymbol}
\usepackage{ dsfont }
\usepackage{balance}

\usepackage{microtype}

\usepackage{enumitem}

\usepackage{graphicx}
\usepackage{caption}

\usepackage{esvect}

\usepackage{anyfontsize}
-lmtk10

\usepackage{color}

\newcommand{\smallsim}{\smallsym{\mathrel}{\sim}}
\makeatletter
\newcommand{\smallsym}[2]{#1{\mathpalette\make@small@sym{#2}}}
\newcommand{\make@small@sym}[2]{%
  \vcenter{\hbox{$\m@th\downgrade@style#1#2$}}%
}
\newcommand{\downgrade@style}[1]{%
  \ifx#1\displaystyle\scriptstyle\else
    \ifx#1\textstyle\scriptstyle\else
      \scriptscriptstyle
  \fi\fi
}
\makeatother

\makeatletter
  \newcommand\smallernormal{\@setfontsize\smallernormal{9.5pt}{9.5}}
\makeatother
\makeatletter
  \newcommand\smallernormalAppendix{\@setfontsize\smallernormalAppendix{11pt}{10}}
\makeatother

\usepackage[hang,flushmargin]{footmisc}
\title{A Disentangled Adversarial Neural Topic Model\\ for Separating Opinions from Plots in User Reviews}

\author{ Gabriele Pergola, Lin Gui, Yulan He \\
 Department of Computer Science, University of Warwick, UK \\
  \texttt{\{gabriele.pergola, lin.gui, yulan.he\}@warwick.ac.uk} \\}

\begin{document}
\maketitle
\begin{abstract}
The flexibility of the inference process in Variational Autoencoders (VAEs) has recently led to revising traditional probabilistic topic models giving rise to Neural Topic Models (NTMs). Although these approaches have achieved significant results, surprisingly very little work has been done on how to disentangle the latent topics. Existing topic models when applied to reviews may extract topics associated with writers' subjective opinions mixed with those related to factual descriptions such as plot summaries in movie and book reviews. It is thus desirable to automatically separate opinion topics from plot/neutral ones enabling a better interpretability. 
In this paper, we propose a neural topic model combined with adversarial training to disentangle opinion topics from plot and neutral ones.
We conduct an extensive experimental assessment introducing a new collection of movie and book reviews paired with their plots, namely \textsc{MOBO} dataset, showing an improved coherence and variety of topics, a consistent disentanglement rate, and sentiment classification performance superior to other supervised topic models.
\end{abstract}

%%%%%%%%%%%%%%%%%%%%%%
% INTRODUCTION
%%%%%%%%%%%%%%%%%%%%%%
\section{Introduction}

Variational Autoencoders (VAEs) \cite{KingmaW13} allow to design complex generative models of data. 
% since the inference process of VAE-based approaches has the advantage of being independent from the model architecture providing high flexibility in designing new neural components.
In the wake of the renewed interest for VAEs, traditional probabilistic topic models \cite{Blei03} have been revised giving rise to several Neural Topic Model (NTM) variants, such as NVDM \cite{MiaoYB16},  ProdLDA \cite{Srivastava17}, and NTM-R \cite{Ding18}. Although these approaches have achieved significant results via the neural inference process, existing topic models when applied to user reviews may extract topics with writers' subjective opinions mixed with those related to factual descriptions such as plot summaries of movies and books \cite{Lin12}. Yet surprisingly very little work has been done on how to disentangle the inferred topic representations. 
% GSM \cite{miao17}, W-LDA \cite{Nan19}

\begin{figure}[!t]
\centering
\includegraphics[width=0.385\textwidth]{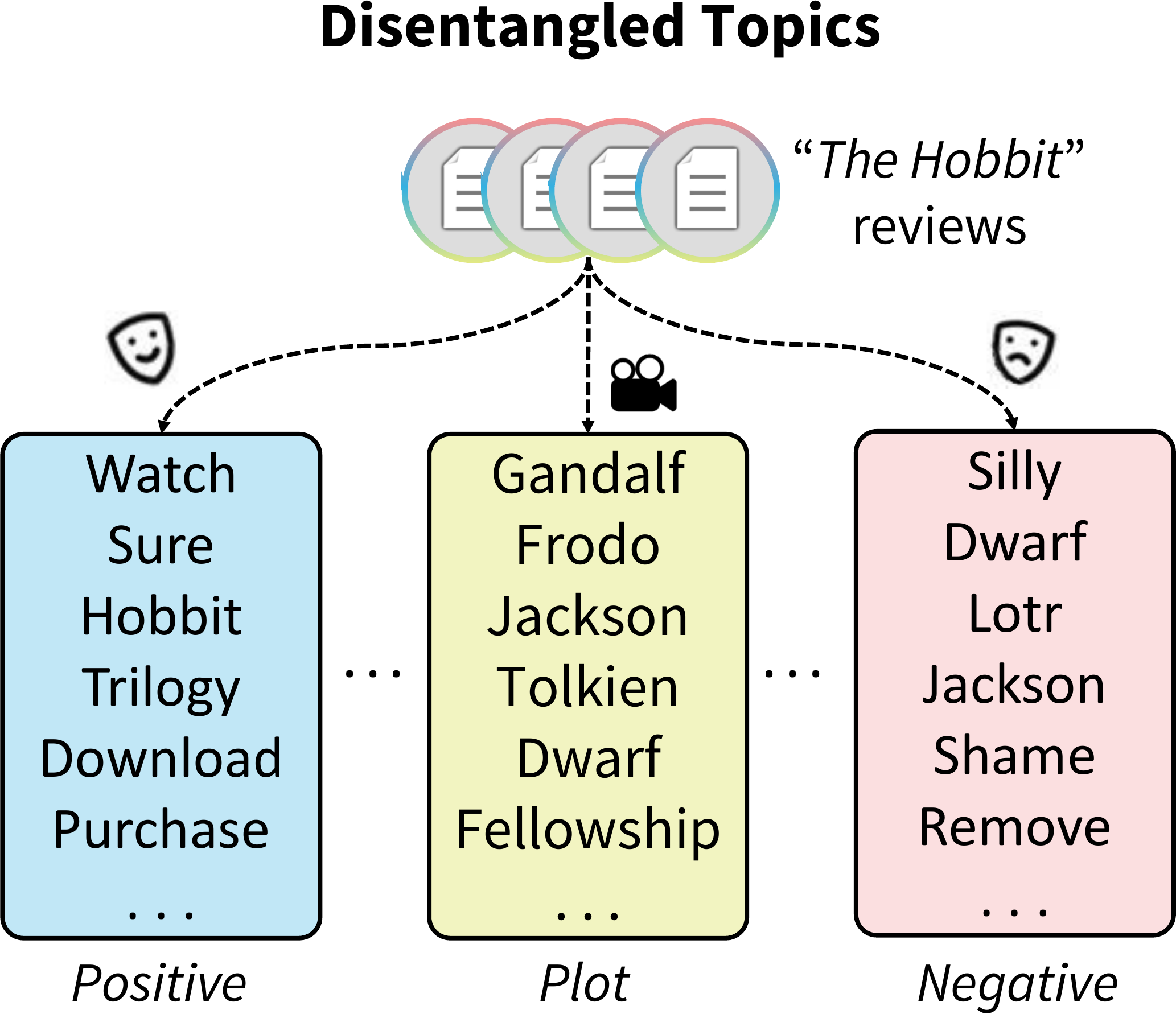}
\caption{Disentangled topics extracted by DIATOM from the Amazon reviews for ``The Hobbit''.}
\vspace{-12pt}
\label{fig:ex_diatom}
\end{figure}

Disentangled representations can be defined as representations where individual latent units are sensitive to variations of a single generative factor, while being relatively invariant to changes of other factors \cite{Bengio13, Higgins17}. Inducing such representations has been shown significantly beneficial for their generalization and interpretability \cite{Achille18, Peng19}. 
%For example, an image can be viewed as the results of several generative factors mutually interacting, as one or many sources of light, the material and reflective properties of various surfaces or the shape of the objects depicted \cite{Bengio13}.
% 
In the context of topic modeling, %documents result from a generative process over mixtures of latent topics, and therefore, 
we propose to consider latent topics as generative factors to be disentangled to improve their interpretability and discriminative power. Disentangled topics are topics invariant to the factors of variation of text, which for instance, in the context of book and movie reviews could be the author's opinion (e.g. positive/negative), the salient parts of a plot or other auxiliary information reported. An illustration of this is shown in Figure~\ref{fig:ex_diatom} in which opinion topics are separated from plot topics.

% where this leads to separating topics based on the ``factor of variation" they are revealing.
% For example, in generating a book review, the factors of variation involved could depend on the author's expertise in identifying the salient features of the book, %his knowledge of the book's genre, or 
% his ability to summarize the plot and the feelings evoked by the book. 
% % [Let's break in/the atom]
% % Figure \ref{ex_diatom} reports a examples of polarity-disentangled topics generated from the IMDB movie reviews of "The Hobbit". The topics on the left and right summarize some of the positive and negative aspects described by users, while neutral topics in the middle report the main elements of the movie's plot.

% An effective approach for disentangling features in the latent space of VAEs is to adopt adversarial training \cite{Mathieu16}. However, despite its successful applications in computer vision \cite{LiuNIPS18}, the applications to text analysis has been rather limited so far \cite{Rui19,Tomonari18}, narrowed by the lack of proper tasks to evaluate the generated disentangled representations and the limited availability of suitable datasets.

% For example, in book or movie reviews, we want to disentangle topics which are related to opinions expressed in text and topics relating to book/movie plots. An illustration of this is shown in Figure~\ref{fig:ex_diatom} in which opinion topics are separated from plot topics. 

However, models relying solely on sentiment information are easily misled and not suitable to disentangle opinion from plots, since even plot descriptions frequently make large use of sentiment expressions \cite{pan04}. Consider for example the following sentence: ``The ring holds a \textit{dark} power, and it soon begins to exert its \textit{evil influence} on Bilbo'', an excerpt from a strong positive Amazon's review.

% This overcomes the difficulty of separating opinions from plot and auxiliary information yet containing polarised descriptions that easily mislead models merely relying on sentiment lexicon; analogously to the issue of mixed topics generated when traditional topic models are applied to review documents, as pointed out in \citet{Blei08}. 
% Despite its successful employment in computer vision \cite{LiuNIPS18}, the adversarial approach has had a rather limited application in text analysis so far \cite{Rui19,Tomonari18}, narrowed by the lack of proper tasks to evaluate the generated disentangled representations and the limited availability of suitable datasets.

Therefore, we propose to distinguish opinion-bearing topics from plot/neutral ones combining a neural topic model architecture with adversarial training. In this study, we present the DIsentangled Adversarial TOpic Model (\textsc{DIATOM})\footnote{\label{ft:diatom_mobo}Source code and dataset available at: \url{https://github.com/gabrer/diatom}.}, aiming at disentangling information related to the target labels (i.e. the review score), from other distinct aspects yet possibly still polarised (e.g. plot descriptions).
We also introduce a new dataset, namely the \textsc{MOBO} dataset\footnotemark[\value{footnote}], made up of movie and book reviews, paired with their related plots. The reviews come from different publicly available datasets: IMDB \cite{Maas11}, GoodReads \cite{Wan19} and Amazon reviews \cite{McAuley15}, %\cite{Maas11, Kar18, Wan19, McAuley15}, 
and encompass a wide spectrum of domains and styles.
We conduct an extensive experimental assessment of our model. First, we assess the topic quality in terms of topic coherence and diversity and compare DIATOM with other supervised topic models on the sentiment classification task; then, we analyse the disentangling rate of topics to quantitatively assess the degree of separation between actual opinion and plot/neutral topics.  

\noindent Our contributions are summarized below:
\begin{itemize}[topsep=0pt]
\itemsep 0.05em 
\item We propose a new model, DIATOM, which is able to generate disentangled topics through the combination of VAE and adversarial learning;
\item  We introduce the \textsc{MOBO} dataset, a new collection of movie and book reviews paired with their plots;
\item 
We conduct an experimental assessment of our model, highlighting more interpretable topics with better topic coherence and diversity scores compared to others state-of-the-art supervised topic models, and improved discriminative power on sentiment classification, and a consistent topic-disentanglement rate.
\end{itemize}

%\noindent The rest of the paper is organized as follows. We review the related literature on sentiment-topic models, neural topic models and the studies on disentangled representations (\textsection \ref{s:related_work}). Then, we present the details of our proposed DIATOM model (\textsection \ref{s:diatom_architecture}), followed by the experimental setup (\textsection \ref{s:exp_setup}) and results (\textsection \ref{s:exp_results}). Finally, we conclude with a summary of the results and suggestions for future works (\textsection \ref{s:conclusion}).

%%%%%%%%%%%%%%%%%%%%%%%%%%%
\section{Related Work}
\label{s:related_work}

Our work is closely related to three lines of research: sentiment-topic models, neural topic models and learning disentangled representations.

\noindent \textbf{Sentiment-Topic Models.}
% NAACL: To save space use this version
% Probabilistic graphical models for topic extraction is a field that has been extensively studied \cite{Rose04, Wang06, Blei06}. In particular, a general-purpose supervised extension which builds on top of LDA \cite{Blei03} is Supervised-LDA (sLDA) \cite{Blei08}, which adds a response variable associated to each document (e.g. a review's rating). 
% 
% NAACL COMMENTED
Probabilistic graphical models for topic extraction have been extensively studied. %a vast literature. 
Beyond LDA \cite{Blei03}, a wide spectrum of models has specialised it to more particular tasks using contextual information \cite{Rose04, Wang06, Blei06, Perg18, Perg19}. Supervised-LDA (sLDA) \cite{Blei08} is a general-purpose supervised extension which builds on top of LDA by adding a response variable associated with each document (e.g. a review's rating). 
A category of extensions particularly relevant for this work is the sentiment-topic models. Examples include the Joint Sentiment-Topic (JST) model \cite{Lin09, Lin12} and Aspect and Sentiment Unification Model (ASUM) \cite{jo2011aspect}. These models are able to extract informative topics grouped under different sentiment classes. Although they do not rely on document labels, they require word prior polarity information to be incorporated into the learning process in order to generate consistent results. 
Nevertheless, The possibility to supervise the learning process with document labels makes JST suitable for a fair comparison.
% 
% 
% NAACL COMMENTED
% Nevertheless, when provided with document-level class labels, JST can learn document-topic distributions influenced by the class information.
% The possibility to supervise the learning process with document labels and avoid the necessity of prior information over words makes it suitable for a fair comparison with the model proposed in this work. 
% Besides, these models require carefully tailored inference algorithms, while the standard Gibbs sampling algorithm used can have a high computational cost when fitting large-scale data, with time and memory scaling linearly with the number of documents, leading researchers to devise more sophisticated approaches to make it scalable \cite{gonzalez11}. 
% 
Compared to DIATOM, the discussed sentiment topic models can only distinguish between \textit{polarity}-bearing topics and neutral ones, remaining strictly aligned to the provided labels. Instead, DIATOM is able to generate opinion-bearing topics and plot topics which may still be polarized but not carrying any user's opinion.
% Compared to these models, DIATOM not only generates polarised topics but is able to separate them from polarised or neutral topics (e.g. plots) which however are not related to the target label (i.e. review's score).

\noindent \textbf{Neural Topic Models.}
Neural models provide a more generic and extendable alternative to topic modeling, and therefore, have recently gained increasing interest.  Some of them use belief networks \cite{Mnih14}, or enforce the Dirichlet prior on the document-topic distribution via Wasserstein Autoencoders \cite{Nan19}. Others adopt continuous representations to capture long-term dependencies or preserve word order via sequence-to-sequence VAE \cite{Dieng17, Xu16, Bowman16, Yang17} whose time complexity and difficulty of training, however, have limited their applications. 
Neural Variational Document Model (NVDM) \cite{MiaoYB16} is a direct extension of VAE used for topic detection in text. In NVDM, the prior of latent topics is assumed to be a Gaussian distribution. This is not ideal since it cannot mimic the simplex in the latent topic space. To address this problem, LDA-VAE \cite{Srivastava17} instead used the logistic normal distribution to approximate the Dirichlet distribution. ProdLDA \cite{Srivastava17} extended LDA-VAE by replacing the mixture model of LDA with a product of experts. % which employs a Laplace approximation to make the gradient back-propagate to the variational distribution. 
\textsc{Scholar} is a neural framework for topic models with metadata incorporation \citep{Card18}. %without the need of deriving model-specific inference  
When metadata are document labels, the model infers topics that are relevant to those labels.
Although some studies have applied the adversarial approach \cite{Goodfellow14} and reinforcement learning \cite{Gui19} to topic models setting a Dirichlet prior on the generative network \cite{Rui19, Tomonari18}, it is still unexplored how to use this mechanism to disentangle opinion-bearing topics from plot or neutral topics. %high-level topic characteristics, as for instance, the carried sentiment polarity.
% Since W-LDA is not based on variational inference, we cannot compute the ELBO based perplexity as a performance metric as in. 

% NAAACL Commented
% Compared to these neural topic models, DIATOM is the first attempt using an adversarial mechanism to distinguish between topic type (i.e. opinion and plot topics), while not only generating topics aligned with the available target labels (i.e. opinion topics) but seamless incorporating the external signal of plot summaries to drive the generation of topics about salient parts of plots mentioned by users (i.e. plot topics) not related to the target classes (i.e. sentiment polarity).

\noindent \textbf{Representation Disentanglement.}
% NAAACL Commented
% Despite the lack of general consensus about a unique definition of disentangled representations \cite{Higgins18, Gao19}, it typically refers to representations which are only sensitive to one single generative factor of data and relatively invariant to other factors of variation \cite{Bengio13}. 
% One proposed definition builds upon the concept of statistical independence by minimizing total correlation \cite{DinhKB14, Achille18}, while an alternative approach explored the possibility to measure and track the changes in a single latent dimension as degree of disentanglement \cite{Higgins17}.
% However, 
Among the slightly different versions of representation disentanglement proposed \cite{Bengio13, Higgins18, Gao19}, the one achieved in DIATOM is analogous to \citet{Thomas17} and \citet{Bengio17}, where they impose additional constraints to the representations controlled using a reinforcement learning mechanism determining the disentangled factors. Alternatively, in DIATOM we  make use of an adversarial approach over the available target labels.
% 
% NAACL COMMENTED
% Both Generative Adversarial Networks (GAN) \cite{Mathieu16, Makhzani15, Chen16, LiuNIPS18} and VAEs \cite{Peng19, Chen18, Hsieh18} have been successfully employed in disentangling features in computer vision tasks. 
Application in text processing has shown promising results \cite{Vineet19, Kumar2017, Hoang2019, Esmaeili19}, yet applications to topic modeling are still limited \cite{Wilson16} and to the best of our knowledge, there is no work in separating opinion-bearing topics from plot/neutral topics.

%%%%%%%%%%%%%%%%%%%%%%
% ARCHITECTURE
%%%%%%%%%%%%%%%%%%%%%%
\section{DIATOM architecture}
\label{s:diatom_architecture}

\begin{figure}[!t]
\centering
\includegraphics[width=1.01\columnwidth]{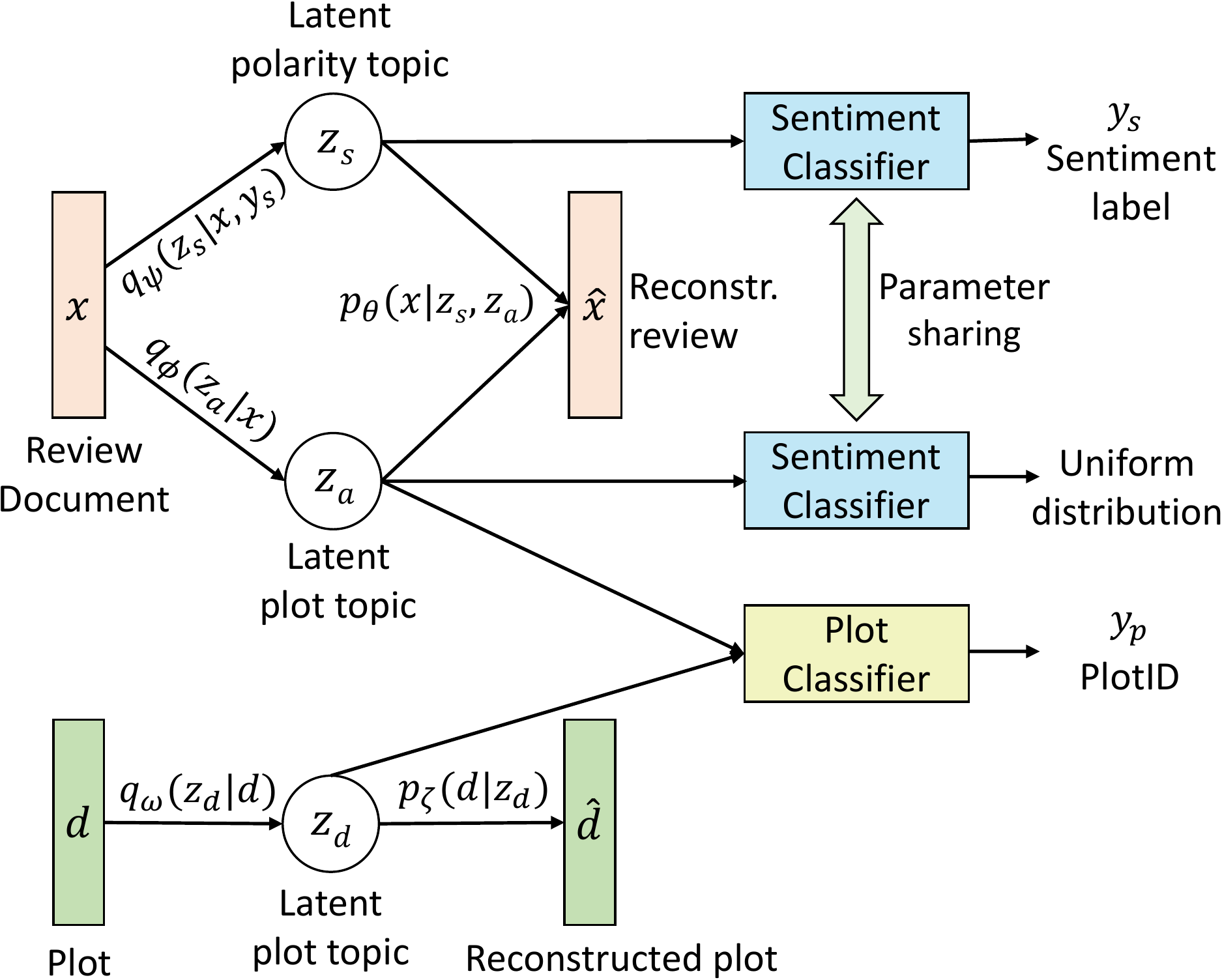}
\caption{The \textsc{DIATOM} Architecture.} 
\label{fig:diatom_arch}
\end{figure}
% \vspace{-12pt}

Our proposed DIATOM model is shown in Figure~\ref{fig:diatom_arch}.
Assuming a document $\bs{x}$ is associated with a sentiment label $y_s$, and each document can be represented by latent topics associated with sentiments ($\bs{z}_s$) and plots\footnote{These are the topics not associated with the target sentiments, which can be either plot topics or neutral topics (not about plots). For notational convenience, we call both plot topics.} ($\bs{z}_a$), we aim to learn a model maximizing the joint data-label log-likelihood, $\log p(\bs{x},y_s)$:
\vspace{-10pt}
\begin{align}\small
&\log p(\bs{x},y_s) = \log\int\int p(\bs{x}, y_s, \bs{z}_a, \bs{z}_s) d\bs{z}_a d\bs{z}_s  \nonumber\\
& ~~~~\ge \mathbb{E}_{q_{\phi}(\bs{z}_a|\bs{x}), q_{\psi}(\bs{z}_s|\bs{x},y_s)} [\log p_{\theta}(\bs{x} | \bs{z}_a, \bs{z}_s) ] \nonumber \\ 
& ~~~~+\mathbb{E}_{q_{\phi}(\bs{z}_a|\bs{x}), q_{\psi}(\bs{z}_s|\bs{x},y_s)} [\log p_{\pi}(y_s | \bs{x}) ] \nonumber\\ 
& ~~~~-\operatorname{KL}\big(q_{\phi}(\bs{z}_a | \bs{x} ) || p(\bs{z}_a ) \big) \nonumber \\
& ~~~~- \operatorname{KL}\big(q_{\psi}(\bs{z}_s | \bs{x},y_s ) || p(\bs{z}_s) \big)   
\label{eq:ELBO}
\end{align}
% \vspace{-26pt}

\noindent Inspired by \citet{MiaoYB16} and \citet{Card18}, we assume the document-level topic distribution for plots can be approximated by a multi-layer perceptron (MLP) taking as input a multivariate Gaussian distribution, and similarly for the topic distribution for sentiments. The multinomial distribution over words under a plot topic and an opinion topic can be parametrised by a weight matrix $\bs{W}$. The generative process is shown below. 
\begin{itemize}
\item For each document $d\in\{1,..,D\}$,
	\begin{itemize}
		\item Draw the latent plot-topics, \\ 
		$\hat{\phi} \sim \mathcal{N}(\mu_\phi, \Sigma_\phi)$, $\;\;\, \bs{z}_a = f_{\hat{\phi}}(\hat{\phi}) $
		\item Draw the latent opinion-topics, \\
		\quad $\hat{\psi} \sim \mathcal{N}(\mu_\psi, \Sigma_\psi)$, $\;\; \bs{z}_s = f_{\hat{\psi}}(\hat{\psi}) $
		\item For each word $n\in\{1,..,N_{d}\}$ in $d$,
         	\begin{itemize}
			\item Draw $x_{d,n} \sim p(x_{d,n} | \bs{W}, \bs{z}_a, \bs{z}_s)$
		    \end{itemize}
		    \item Generate the document-level sentiment label, $y_s \sim p(y_s|f_y (\bs{z}_s))$
    \end{itemize}
\end{itemize}
where $f_{\hat{\phi}}$, $f_{\hat{\psi}}$ and $f_y$ are MLPs, $\bs{z}_a$ is a $K$-dimensional latent topic representation of plots for document $d$, $\bs{z}_s$ is a $S$-dimensional latent topic representation of sentiments for document $d$. The probability of word $x_{d,n}$ can be parametrised by another network:
\begin{equation}
\label{eq:wordProb}
p(x_{d,n} | \bs{W}, \bm{z}_a, \bm{z}_s) \propto  \exp\big(\bm{m}_d    %\nonumber
%& + \bm{W}_a \cdot \bs{z}_a + \bm{W}_s \cdot \bs{z}_s \\   
 + \bm{W}\cdot(\bs{z}_a \mathbin\Vert \bs{z}_s)\big)  
\end{equation}
%\vspace{-30pt}

\noindent where $\bs{m}_d$ is the $V$-dimensional background log-frequency word distribution, %$\bm{W}_a\in \mathbb{R}^{V\times K}, \bm{W}_s\in \mathbb{R}^{V\times S},$ 
and $\bm{W}\in \mathbb{R}^{V\times (K+S)}$, while $\bs{z}_a \mathbin\Vert \bs{z}_s$ is the concatenation of the two latent topic vectors.\\

\vspace{-10pt}
%%%%%%%%%%%%%%%%%
\noindent \textbf{Plot Inference Network.} Following the idea of VAE which computes a variational approximation to an intractable posterior using MLPs, we define two inference networks $f_{\mu_{\phi}}$ and $f_{\Sigma_{\phi}}$ which takes as input the word counts in documents:
\begin{equation}
\begin{matrix}
\mu_{\phi} = f_{\mu_{\phi}}(\bs{x}) &
\Sigma_{\phi} = \mbox{diag}(f_{\Sigma_{\phi}}(\bs{x}))
\end{matrix}
\label{eq:aspect_mlp}
\end{equation}
The outputs of both networks are vectors in $\mathbb{R}^K$. Here, `diag' converts a column vector to a diagonal matrix. For a document $\bs{x}$, $q(\phi) \simeq \mathcal{LN}(\mu_{\phi}, \Sigma_{\phi})$. With such a formulation, we can generate samples from $q(\phi)$ by first sampling $\epsilon \sim \mathcal{N}(0, I)$ and then computing $\hat{\phi} = \sigma(\mu_{\phi} + {\Sigma_{\phi}}^{1/2}\epsilon)$. \\

\vspace{-10pt}
%%%%%%%%%%%%%%%%%
\noindent \textbf{Sentiment Inference Network.} Similarly, to compute a variational approximation to $q(\psi)$, we define two inference networks $f_{\mu_{\psi}}$ and $f_{\Sigma_{\psi}}$ which takes as input the word counts in documents:
\begin{equation}
\begin{matrix}
\mu_{\psi} = f_{\mu_{\psi}}(\bs{x}) & 
\Sigma_{\psi} = \mbox{diag}(f_{\Sigma_{\psi}}(\bs{x}))
\end{matrix}
\label{eq:sent_mlp}
\end{equation}
The outputs of both networks are vectors in $\mathbb{R}^S$. For a document $\bs{x}$, $q(\psi) \simeq \mathcal{LN}(\mu_{\psi}, \Sigma_{\psi})$. We then generate samples from $q(\psi)$ by first sampling $\epsilon \sim \mathcal{N}(0, I)$ and then computing $\hat{\psi} = \sigma(\mu_{\psi} + {\Sigma_{\psi}}^{1/2}\epsilon)$. \\

\vspace{-8pt}
%%%%%%%%%%%%%%%%%
\noindent \textbf{Overall Objective.}  With the sampled $\hat{\phi}$ and $\hat{\psi}$, for each document $\bs{x}$, we can minimise the reconstruction loss with a Monte Carlo approximation using $L$ independent samples:
\vspace{-4pt}
\begin{align}
\nonumber
\mathcal{L}_{\bs{x}} & \approx \frac{1}{L}\sum_{l=1}^L \sum_{n=1}^{N_d} \log p_{\theta}(x_{d,n} | \hat{\phi}^{(l)}, \hat{\psi}^{(l)})            \\ \nonumber
& -\operatorname{KL}\big(q(\bs{z}_a | \bs{x} ) \; || \; p(\bs{z}_a ) \big)       \\
& - \operatorname{KL}\big(q(\bs{z}_s | \bs{x},y_s ) \; || \; p(\bs{z}_s) \big) \\ \nonumber
\end{align}
\vspace{-30pt}

\noindent where the first term in the RHS is given by Eq. (\ref{eq:wordProb}).
It has been previously shown in \citet{KingmaW13}, if a standard multivariate normal prior is placed on the latent variables $\bs{z}_a$ and $\bs{z}_s$, then there is a closed form solution to the KL divergence terms above. 

We assume that the latent topics associated with plots, $\bs{z}_a$, are independent of sentiment classes, and hence, when fed into a sentiment classifier, should generate a uniform sentiment class distribution (similar to adversarial learning). On the contrary, the latent topics associated with sentiments, $\bs{z}_s$, should bear essential information to discriminate between sentiment classes. Therefore, we define the following two objectives for sentiment classification; the former being the expected KL divergence with the uniform distribution $\mathcal{U}$, and the latter a cross-entropy loss:
% \vspace{-10pt}
\begin{align}
\mathcal{L}_{adv} &= -\mathbb{E}_{q_{\phi}(\bs{z}_a)} \, \big[\operatorname{KL}\big( \mathcal{U}(0,M) \; || \; p(\hat{y} | \bs{z}_a) \big)\big] \nonumber\\
\mathcal{L}_{sent} &= -\mathbb{E}_{q_{\psi}(\bs{z}_s)} \sum_{c=1}^M y_c \log\big(p(\hat{y}_c | \bs{z}_s)\big) 
\end{align}
\vspace{-10pt}

\noindent where $M$ is the total number of sentiment classes, and $\mathcal{U}(0,M)$ represents the uniform sentiment class distribution.

To further disentangle the latent topics associated with plots and sentiments,
% $\bs{z}_a$, and latent topics associated with sentiment, $\bs{z}_s$, 
while concurrently minimise the redundancy in the final topic matrix, we apply an orthogonal regularizer over the decoder matrix $\bm{W}$. $\mathcal{L}_{orth}$ reaches its minimum value when the dot product between different topic distributions goes close to zero:

\vspace{-10pt}
\begin{equation}
    \mathcal{L}_{orth} = ||\,\bm{W} \cdot \bm{W}^T - \mathds{I} \, ||
\end{equation}

% mutual information between them:
% \begin{equation}
%     \mathcal{L}_{MI} = \mathbb{E}_{q_{\theta}(\bs{z}_a)q_{\psi}(\bs{z}_s)} \log\frac{p(\bs{z}_a, \bs{z}_s)}{p(\bs{z}_a) p(\bs{z}_s)}
% \end{equation}

\noindent Our final objective function is:
\begin{equation}
    \mathcal{L} =  -\alpha\mathcal{L}_{\bs{x}} + \beta\mathcal{L}_{adv} + \gamma\mathcal{L}_{sent} +  \eta\mathcal{L}_{orth}
    \label{eq:overall_loss}
\end{equation}
\noindent where $\alpha, \beta$, $\gamma$ and $\eta$ control %the signs} and 
the relative contribution of various loss functions. \\

\vspace{-10pt}

%%%%%%%%%%%%%%%%%
\noindent \textbf{Plot Network.} An additional VAE is plugged to the model providing a supplementary signal for the latent plot topic extraction. This mechanism preserves the plot information that might contain some sentiment words and thus, be wrongly regard as a user's opinion. The inference network is defined analogously to Eq. \ref{eq:aspect_mlp}, which instead of taking a review document, takes a plot summary as input. An additional cross-entropy objective is minimized to drive the latent plot topics ($\bs{z}_a$) which would have a similar discriminative power as %with as discriminative as 
the features ($\bs{z}_d$) directly derived from the plots when used for plot classification:
% close to the supplementary information ($\bs{z}_d$):

\vspace{-18pt}
\begin{align}
&\mathcal{L}_{\bs{d}} = \mathbb{E}_{q_{\omega}(\bm{z}_d|d)}[p_{\zeta}(d|\bm{z}_d)] -\operatorname{KL}\big(q(\bs{z}_d | \bs{d} ) \; || \; p(\bs{z}_d ) \big)\nonumber\\
&\mathcal{L}_{plot_{{z}_a}} = -\mathbb{E}_{q_{\phi}(\bs{z}_a)} \sum_{p=1}^P y_p \log(p\big(\hat{y}_p | \bs{z}_a) \big)  \nonumber\\
&\mathcal{L}_{plot_{{z}_d}} = -\mathbb{E}_{q_{\omega}(\bs{z}_d)} \sum_{p=1}^P y_p \log\big(p(\hat{y}_p | \bs{z}_d) \big) 
\end{align}

% \begin{equation}
% \mathcal{L}_{plot} = -\mathbb{E}_{q_{\psi}(\bs{z}_s)} \sum_{d=1}^D y_d \log(p(\hat{y}_d | (\bs{z}_a \mathbin\Vert \bs{z}_d))  
% \end{equation}

\noindent where $P$ denotes the total number of plots in each dataset. Finally, $-\mathcal{L}_{\bm{d}}$ and $\mathcal{L}_{plot}$ are added to the overall loss defined in Eq. \ref{eq:overall_loss}.

%%%%%%%%%%%%%%%%%%%%%%
% Experimental SETUP
%%%%%%%%%%%%%%%%%%%%%%
\section{Experimental Setup}
\label{s:exp_setup}
We conduct thorough experimental evaluations to assess the quality and disentanglement rate of extracted topics. To assess the quality of topics, we compute their topic coherence \cite{Roder15} coupled with their topic uniqueness.
Then, we additionally look at the discriminative power of the disentangled features on the sentiment classification task. 
To fully assess the disentanglement rate of different methods, we perform topic labeling to compute the sentiment polarity of each topic (if any) and then measure the overall disentanglement rate (\textsection \ref{subs:topic_disent}).
% (Eq. \ref{eq:disent_rate_def}). 
As a result, we obtain an estimate of the extent to which different models can accurately control the topic disentanglement rate. 
We introduce and use a new dataset, named the \textit{MOBO} dataset, pairing movie/book plots with their users' reviews, and including human-annotated sentences.
% used to evaluate the disentanglement rate of opinion-bearing topics and plot (or neutral) topics. 

% we measure the sentiment polarity of topics and the disentangling rate among different classes, as polarized (i.e. \texttt{positive} and \texttt{negative}) and neutral (i.e. \texttt{none} and \texttt{plot}). %Metti a small i texttt come nel report
% we compute the \textit{Topic-Purity coherence} (\textsection \ref{s:topic_purity_coh}), and a new dataset, i.e. \textit{MOBO dataset} (\textsection \ref{s:mobo_dataset}).

%%%%%% DATASET %%%%%%
\noindent\textbf{MOBO Dataset}.
\label{s:mobo_dataset}
The MOBO dataset is a collection of reviews and plots about \textbf{MO}vie and \textbf{BO}ok, associated to human-annotated sentences: while the pairs of reviews and plots are used to enhance the generation of plot topics, the human-annotated sentences provide the necessary ground-truth to automatically evaluate the topics' polarity. 

Movie and book reviews were collected and paired from 3 public datasets: the Stanford's IMDB movie reviews \cite{Maas11}, the GoodReads \cite{Wan19} and the Amazon reviews dataset \cite{McAuley15}.
Among all the available reviews in the IMDB dataset, we keep the ones with a corresponding plot in the MPST dataset \cite{Kar18}, a corpus of movie synopses. The Goodreads dataset comes already with books' reviews paired with the related plots; while from the Amazon dataset, among all the product reviews, we keep only the ones related to movies available on the store and whose descriptions consist of the movie plots\footnote{The dataset provides a predefined split of the corpus which preserves on train, development and test sets the same distribution of reviews based on their corresponding plots.}. 
% 
% CAMERA-READY
With the help of 15 annotators we further labeled more than 18,000 reviews' sentences ($\smallsim6000$ per corpus), marking the sentence polarity (\texttt{Positive}, \texttt{Negative}), or whether a sentence describes its corresponding movie/book \texttt{Plot}, or none of the above (\texttt{None})\footnote{We use \textit{Doccano} as framework for collaborative labelling: https://github.com/doccano/}. 
We ensured that each sentence was labelled by at least 2 annotators by assigning overlapping subsets of $\sim2400$ sentences. In case of disagreement, when no consensus was reached, a final choice was made through a majority vote involving a third annotator. The final inter-annotator agreement (Cohen’s kappa) was computed between each pair of annotators sharing a common subsets, with a minimum value of 0.572 and maximum of 0.831, for a resulting average of 0.758\footnote{We publicly release the full set of sentences with and without annotations for future expansion.}.
We report the dataset statistics in Table \ref{tb:dataset_statistics}. We report the dataset statistics in Table \ref{tb:dataset_statistics}. 

\begin{table}[!t]\small
\center
\begin{tabular}{lrrr}
\toprule
\textbf{Statistics} & \textbf{IMDB} & \textbf{GoodReads}  & \textbf{Amazon}\\
\midrule
\# plots       & 1,131        & 150    & 100  \\
\# reviews    & 25,836  & 83,852  & 32,375     \\
%Avg. No. of reviews / plot % (avg / max / min) & 24 / 30 / 10    & 954 / 3,000 / 549    & 464 / 1525 / 272     \\
%& 24    & 954    & 464    \\
%Pos/Neg/Neu distri.  & 0.46/0.54/0  & 0.33/0.50/0.17  & 0.32/0.46/0.22  \\
\% Pos. reviews  & 0.46  & 0.33  & 0.32  \\
\% Neg. reviews  & 0.54  & 0.50  & 0.46  \\
\% Neu. reviews  & 0  & 0.17  & 0.22  \\
Training set  & 20,317 & 65,816   & 25,883   \\
Dev. set  & 2,965    & 9,007    & 3,275   \\
Test set    & 2,554   & 9,029   & 3,217   \\
\midrule
\# annotated sent. & 6,000  & 6,000 & 6,000      \\
%Pos / Neg / Plot / None distribution    &  &  & \\
\bottomrule
\end{tabular}
\caption{The MOBO dataset statistics.}
\vspace{-8pt}
\label{tb:dataset_statistics}
\end{table}

% 
% We removed books or movies with only positive or negative reviews.
% Amazon dataset: from section "Movies_and_TV" removing reviews containing the following words: "disc", "discs", "dvd", "dvds", "vhs", "stereo", "subtitles", "tv", "workout", "documentary", "cofanetto", "ph.d.", "theatre", "guide", "university"

%%%%  TopicPolarity - TopicUniqueness - TopicCoherence (NPMI) %%%%
\begin{table*}[htb]\small
\centering
\setlength{\tabcolsep}{20pt}
% \resizebox{\textwidth}{!}{ 
% \resizebox{1.8\columnwidth}{!}{%
\scalebox{0.9}{\begin{tabular}{@{}l@{}lcccc@{}}
\toprule
\textbf{Datasets} &  \textbf{Models} & \multicolumn{3}{c}{\textbf{Topic Coherence $\,/\,$ Topic Uniqueness }} \\ 
    
    % \begin{tabular}{@{}c@{}}\textit{Polarized-Neutral} \\ \textit{topics} \end{tabular} & $n/a$ & $15-50$ & $50-50$ & $n/a$ & $15-50$ &$50-50$ & $n/a$ & $15-50$ & $50-50$ \\ 
      &  & $25$ & $50$ & $100$ & $200$ \\
    \midrule
    
IMDB  & LDA              & 0.395 $/$ 20.3       & 0.387 $/$ 30.1                           & 0.383 $/$ 33.9             & 0.391 $/$ 34.4         \\
      & sLDA             & 0.421 $/$ 15.8       & 0.376 $/$ 18.9                           & 0.291 $/$ 13.5             & 0.288 $/$ 14.6         \\
      & \textsc{JST}     & 0.472 $/$ 22.7       & 0.526 $/$ 26.8                           & 0.527 $/$ 29.3             & 0.530 $/$ 31.1         \\\cmidrule{2-6}
      & \textsc{NVDM}    & 0.281 $/$ 15.8       & 0.284 $/$ 30.2                           & 0.273 $/$ \textbf{50.3}    & 0.266 $/$ \textbf{54.8}\\
      & \textsc{GSM}     & 0.384 $/$ 22.4       & 0.402 $/$ 21.0                           & 0.410 $/$ 39.7             & 0.394 $/$ 42.4         \\
      & \textsc{NTM}     & 0.423 $/$ 28.8       & 0.508 $/$ 28.6                           & 0.513 $/$ 24.1             & 0.523 $/$ 23.5         \\
      & \textsc{ProdLDA} & 0.502 $/$ 31.1       & 0.543 $/$ 30.8                           & 0.566 $/$ 27.7             & 0.558 $/$ 29.2         \\
      & \textsc{Scholar} & \textbf{0.550} $/$ 28.4     & 0.616 $/$ 27.0                    & 0.618 $/$ 29.7             & 0.593 $/$ 31.5         \\\cmidrule{2-6}
      & \textsc{DIATOM}  & 0.544  $/$ \textbf{37.1}    & \textbf{0.639} $/$ \textbf{38.1}  & \textbf{0.626} $/$ 36.5    & \textbf{0.615} $/$ 30.7 \\
    %   NAACL Commented
    %   & -- w/o Plot Network  & 0.525  $/$ 30.1           & 0.603 $/$ 36.7            & 0.607 $/$ 33.8             & 0.584 $/$ 30.3         \\
      \midrule    \midrule
      
      GoodReads $\quad$  & LDA  & 0.441 $/$ 19.6     & 0.463 $/$ 33.5         & 0.455 $/$ 41.6               & 0.462 $/$ 40.3        \\
      & sLDA             & 0.432 $/$ 34.4              & 0.387 $/$ 47.3         & 0.313 $/$ 25.6             & 0.315 $/$ 23.8        \\
      & \textsc{JST}     & 0.465 $/$ 43.5              & 0.549 $/$ 46.2         & 0.560 $/$ 47.6             & 0.551 $/$ 45.2        \\\cmidrule{2-6}
      & \textsc{NVDM}    & 0.294 $/$ 40.8            & 0.323 $/$ 30.2         & 0.287 $/$ 48.3               & 0.264 $/$ 46.9        \\
      & \textsc{GSM}     & 0.411 $/$ 24.8            & 0.481 $/$ 40.1         & 0.482 $/$ 38.1               & 0.473 $/$ 41.4        \\
      & \textsc{NTM}     & 0.421 $/$ 23.5            & 0.523 $/$ 47.6         & 0.493 $/$ 33.4               & 0.465 $/$ 38.7        \\
      & \textsc{ProdLDA} & 0.551 $/$ 30.3            & 0.562 $/$ 41.8         & 0.564 $/$ 39.8               & 0.556 $/$ 37.7        \\
      & \textsc{Scholar} & 0.545 $/$ 38.3              & 0.603 $/$ 42.0         & \textbf{0.681} $/$ 41.2    & \textbf{0.664} $/$ 38.4        \\\cmidrule{2-6} 
      & \textsc{DIATOM}  & \textbf{0.582} $/$ \textbf{54.0}  & \textbf{0.634} $/$ \textbf{52.9}  & 0.628 $/$ \textbf{54.9}  & 0.609 $/$ \textbf{48.7}        \\
    %   & -- w/o Plot Network  & 0.555  $/$ 40.1           & 0.615 $/$ 49.3                  & 0.607 $/$ 33.8           & 0.578 $/$ 39.6        \\
      \midrule
      \midrule
      
      Amazon & LDA       & 0.430 $/$ 28.9          & 0.447 $/$ 47.5           & 0.438 $/$ 64.8      & 0.445 $/$ 59.3    \\
      & sLDA             & 0.421 $/$ 67.7          & 0.393 $/$ 62.1           & 0.323 $/$ 87.5      & 0.331 $/$ 74.8     \\
      & \textsc{JST}     & 0.450 $/$ 73.0          & 0.558 $/$ 71.2           & 0.544 $/$ 78.8      & 0.518 $/$ 70.9     \\\cmidrule{2-6}
      & \textsc{NVDM}    & 0.278 $/$ 42.4          & 0.310 $/$ 32.5           & 0.281 $/$ 38.4      & 0.261 $/$ 49.1      \\
      & \textsc{GSM}     & 0.441 $/$ 53.2          & 0.451 $/$ 60.0           & 0.433 $/$ 61.7      & 0.427 $/$ 64.4      \\
      & \textsc{NTM}     & 0.493 $/$ 52.8          & 0.501 $/$ 53.1           & 0.547 $/$ 55.3      & 0.508 $/$ 59.3      \\
      & \textsc{ProdLDA} & 0.492 $/$ 63.4          & 0.543 $/$ 51.4           & 0.528 $/$ 58.7      & 0.551 $/$ 62.1      \\
      & \textsc{Scholar} & 0.548 $/$ 60.5          & 0.587 $/$ 65.1           & \textbf{0.641} $/$ 63.2   & 0.629 $/$ 68.2      \\\cmidrule{2-6}
      & \textsc{DIATOM}  & \textbf{0.563} $/$ \textbf{82.0}  & \textbf{0.598} $/$ \textbf{82.3}  & 0.626 $/$ \textbf{80.8}  & \textbf{0.636} $/$ \textbf{78.5}      \\
    %   & -- w/o Plot Network  & 0.539  $/$ 30.1         & 0.584 $/$ 78.3                    & 0.611 $/$ 73.4     & 0.618 $/$ 74.7         \\
    %   &  -- w/o orthogonal reg.  & 0.544  $/$ 37.1   	& 0.639 $/$ 38.1       & 0.626 $/$ 36.5\\

    \bottomrule
  \end{tabular}
}
% }
\caption{Topic Coherence and Topic Uniqueness results for 25/50/100/200 topics. The best result in each column and for each dataset is highlighted in \textbf{bold}.}
\vspace{-10pt}
\label{tb:aspect_coherence}
\end{table*}

%%%%%%%%%%%%%%%%%%%%%%%%%%%%%%%
%\noindent\textbf{Evaluation Metrics}.
% \noindent \textbf{Evaluation.} 
%For a single topic, the topic polarity is identified by first matching a topic $k_i$ with $S$ human-annotated sentences; then, the most frequent label among these sentences is assumed as the topic's polarity. Topics and sentences are matched based on the cosine similarity between the respective embeddings. The topic embedding results from the normalized weighted average of the top-N words vectors generate by BERT: $\vv{t_{z}} = \frac{1}{N}\sum_{i=1}^N \alpha_i*\vv{w_i}$, where $alpha_v$ is the word weights in the specific topic-word distribution. The sentence embedding $\vv{s_j}$ is computed using the Sentence-BERT encoder \cite{reimers2019}, an off-the-shelf encoder which has been proven to generate meaningfull cost-effective representations for sentences. Finally, topics and sentences are paired picking the 10 sentences with the highest cosine similarity score for each topic and the most frequent labels among the sentences is adopted as the topic's polarity.

\noindent\textbf{Baselines}.
\label{ss:baselines}
We compare the experimental results with the following baselines:

\noindent\underline{sLDA} \cite{Blei08}: a supervised extension of LDA adding a response variable associated with each document.

\noindent\underline{JST} \cite{Lin09}: Joint Sentiment-Topic model built on LDA which is able to extract polarity-bearing topics.

\noindent\underline{NVDM} \cite{MiaoYB16}: a VAE with an encoder network mapping the bag-of-words representations into continuous latent distributions, and a generative network for the document reconstruction.

\noindent\underline{GSM} \cite{miao17}: based upon NVDM, the Gaussian Softmax topic model generating the topic distribution by applying a softmax function on the hidden representations of documents.

\noindent\underline{NTM} \cite{Ding18}: Neural Topic Model is a variation of NVDM by plugging the topic coherence metric directly into the model's objective.

\noindent\underline{\textsc{ProdLDA}} \cite{Srivastava17}: 
% with an architecture similar to NVDM, 
ProdLDA introduces a Dirichlet prior in place of Gaussian prior over the latent topic variable.

\noindent\underline{\textsc{Scholar}} \cite{Card18}: a neural framework based on variational inference for the generation of topic while incorporating metadata information.

%\noindent For a fair comparison, we only compare the supervised models in terms of sentiment classification accuracy.

%%%%%%%%%%%%%%%%%%%%%%%%%%%%%%%
\noindent\textbf{Parameter Setting}.
We perform tokenization and sentence splitting with SpaCy\footnote{https://spacy.io/}.
When available, we keep the default preprocessing, as it is the case for sLDA and \textsc{Scholar}. Along with stopwords, we also remove tokens shorter than three characters and those with just digits or punctuation. We set the vocabulary to the 2,000 most common words as the best trade-off for each dataset. 
The 300-dimensional word vectors are initialized with a pretrained BERT embedding \cite{devlin19}. Sentence embeddings are generated from the Sentence-BERT using a pretrained BERT-large with mean-tokens pooling \cite{reimers2019}.
% \footnote{After preliminary experiments with FastText \cite{bojanowski17} and Glove \cite{Pennington14}). 
% We vary the dictionary size depending on the dataset at hand, as described in Table \ref{tb:dataset_statistics}.
We use the predefined split of the MOBO dataset into training, development and test set in the proportion of 80/10/10 and average all the results over 5 runs.\footnote{Hyperparameter setting and training details are in Appendix.}

\section{Experimental Results}
\label{s:exp_results}

We report the results in terms of topic coherence/uniqueness, sentiment classification and topic disentanglement rate. We also perform ablation studies to gain more insights into our model.

% CAMERA-READY
\begin{table*}[htb]\small
\centering
\setlength{\tabcolsep}{15pt}
% \resizebox{\textwidth}{!}{ 
% \resizebox{\columnwidth}{!}{%
%\begin{tabular}{@{}l@{}l@{}c@{}}
\begin{tabular}{lccc}
\toprule
 \textbf{Models} & \textbf{IMDB} & \textbf{GoodReads} & \textbf{Amazon}\\ 
    %   &  & \textit{Held-out set} \\
    \midrule
  SVM    \\  
      % \gab{$\quad$ \text{\footnotesize + BoW}}          & $0.657$       \text{\footnotesize $\pm\,0.01$}  & $0.685$    \text{\footnotesize $\pm\,0.01$}  & $0.653$   \text{\footnotesize $\pm\,0.02$}  \\
       $\quad$ \text{\footnotesize + TFIDF}       & $0.672$       \text{\footnotesize $\pm\,0.02$}  & $0.711$    \text{\footnotesize $\pm\,0.01$}  & $0.661$   \text{\footnotesize $\pm\,0.02$}  \\
       $\quad$ \text{\footnotesize + TFIDF + Lexicon $\qquad$} & $0.683$    \text{\footnotesize $\pm\,0.02$} & $\textbf{0.719}$   \text{\footnotesize $\pm\,0.02$}  & $0.667$   \text{\footnotesize $\pm\,0.02$}   \\
       $\quad$ \text{\footnotesize + LDA}         & $0.615$       \text{\footnotesize $\pm\,0.02$} & $0.659$    \text{\footnotesize $\pm\,0.02$}  & $0.594$   \text{\footnotesize $\pm\,0.01$}   \\ \midrule
       sLDA                & $0.637$           \text{\footnotesize $\pm\,0.01$}     & $0.652$    \text{\footnotesize $\pm\,0.01$} & $0.579$           \text{\footnotesize $\pm\,0.01$}  \\
       \textsc{JST}        & $0.639$           \text{\footnotesize $\pm\,0.01$}     & $0.518$    \text{\footnotesize $\pm\,0.01$}  & $0.538$           \text{\footnotesize $\pm\,0.01$} \\ 
       \textsc{Scholar}    & $0.645$           \text{\footnotesize $\pm\,0.02$}    & $0.673$    \text{\footnotesize $\pm\,0.03$}  & $0.613$           \text{\footnotesize $\pm\,0.02$}  \\ \midrule 
       \textsc{DIATOM}     & $0.726$           \text{\footnotesize $\pm\,0.03$}  & $0.704$    \text{\footnotesize $\pm\,0.02$} & $\textbf{0.686}$  \text{\footnotesize $\pm\,0.02$} \\
       $\quad$ \text{\footnotesize -- w/o Plot Network} & $\textbf{0.734}$  \text{\footnotesize $\pm\,0.03$}  & $0.695$    \text{\footnotesize $\pm\,0.03$} & $0.603$ \text{\footnotesize $\pm\,0.02$}  \\
    \bottomrule
  \end{tabular}
%   }
% }
\caption{Sentiment classification accuracy with 50 topics over the test set.}
\label{tb:sentiment_classification}
\end{table*}

\subsection{Topic Coherence and Uniqueness}
We conduct thorough experimental evaluations to assess the quality and disentanglement rate of extracted topics. To assess the quality of topics, we compute their topic coherence \cite{Roder15} coupled with their topic uniqueness. 
% Traditionally, topic models have been evaluated through the perplexity over held-out documents \cite{Wallach09}.
% Lower perplexity implied better predictiveness as it aimed at measuring the model goodness-of-fit over a held-out set. 
% However, it has been shown that better perplexity does not imply more comprehensible topics \cite{Chang09, Card18} . That is why topic coherence was introduced \cite{lau14}, to evaluate topics regarding their understandability with a score closely matching human judgments.
% Normalized Pointwise Mutual Information (NPMI) is a topic coherence score shown effective in matching the human judgments \cite{Roder15}, and measures the statistical independence of observing two words in close proximity based on the word co-occurrence statistics.
% The NPMI for a list of words $\bm{w}$ is defined in Eq. \ref{eq:npmi_def}:
% % 
% \begin{equation}
% \text{NPMI($\bm{w}$) = } \frac{1}{\text{\textit{N(N-1)}}} \sum_{j=2}^{N} \sum_{i=1}^{j-1} \frac{\log \frac{P\left(w_{i}, w_{j}\right)}{P\left(w_{i}\right) P\left(w_{j}\right)}}{-\text{log} P\text{\textit{(}}w_{i}, w_{j}\text{\textit{)}}}
% \label{eq:npmi_def}
% \end{equation}
% % 
% where $P(w_i)$ and $P(w_i, w_j)$ are calculated based on the word co-occurrences in a reference dataset, and $N$ is typically set to 10, thus considering the top-10 words of topics. The aforementioned definition normalizes the PMI in $[-1,1]$, so that for two words, -1 denotes no co-occurrences while +1 a complete co-occurrence.
We evaluate the topic coherence using the $C_V$ metric, a slightly refined Normalized Pointwise Mutual Information (NPMI) score using a boolean sliding window to determine the words' context \cite{Roder15}.

% The traditionally adopted metrics to evaluate topic models have been perplexity \cite{Wallach09} and topic coherence\cite{Roder15}. 
% % and is defined as the reciprocal geometric mean of the likelihood of a test corpus; 

%Moreover, neural topic models might suffer of topic redundancy, therefore, 
Additionally, we monitor the topic uniqueness (TU) to measure word redundancy across topics. Following \citet{Nan19}, we use $cnt(l,k)$ to denote the total number of times the top word $l$ in topic $k$ appears among the top words across all topics, then $TU(k) = \tfrac{1}{L} \sum_{l=1}^{L} \tfrac{1}{cnt(l,k)}$.
TU is inversely proportional to the number of times each word appears in topics; a higher TU score implies that the top words are rarely repeated and, therefore, more diverse and unique topics.

In Table \ref{tb:aspect_coherence}, we report the topic coherence and the topic uniqueness values. The supervised document label information was incorporated into sLDA, JST, \textsc{Scholar} and DIATOM. Other models are purely unsupervised. We can observe that among conventional LDA-based models, JST performs significantly better compared to both LDA and sLDA for different topic settings and across all datasets. Neural topic models give mixed results. In terms of topic coherence, the trend is \textsc{Scholar} $>$ \textsc{ProdLDA} $>$ NTM $>$ GSM $>$ NDVM. However, when we examine the topic uniqueness values, we can see that higher topic coherence values do not necessarily lead to higher topic uniqueness values. %One example is the results generated by \textsc{Scholar} for the 100 topic setting on the IMDB dataset in which the topic coherence value is high but the uniqueness value is low. 
This shows that the topic coherence value could be misleading sometimes since %The combination of these two metrics shows how some of the neural topic model could, at times, provide 
a high topic coherence could be due to the redundancy of words across topics. We also notice that models with supervised document label information (except sLDA) generally outperform the unsupervised ones. This shows that the document label information can indeed help to extract more meaningful topics. When compared our proposed \textsc{DIATOM} with the baselines, we can observe that it achieves better coherence and topic uniqueness values most of the time, %in 6 out of 9 settings and better topic uniqueness values in 8 out of 9 settings. 
showing the benefit of separating opinion-bearing topics from plot topics by adversarial learning. 
% NAACL COMMENTED
% The importance of the plot network is evident from the results since removing the plot network (``$-$w/o Plot Network'') leads to the degraded topic coherence and topic uniqueness measures.

% CAMERA-READY
%%%%%%%%%%%%%%%%%%%%%%%%%% 
% Sentiment classification
%%%%%%%%%%%%%%%%%%%%%%%%%%
\subsection{Sentiment Classification}

In this section, we compare DIATOM with other supervised topic models for sentiment classification. The purpose of this evaluation is to highlight the discriminative power of the generated representations for the labels of interest while having attractive and unique properties as topic models, rather than confronting them with current state-of-the-art for text classification.
We additionally report some baseline results using a Support Vector Machine (SVM) which has been widely employed on these task \cite{pang04} providing an understanding of the relative differences in performance of different approaches.

Table \ref{tb:sentiment_classification} shows the sentiment classification accuracy. In JST, the supervised document label information is only incorporated as prior to the model, while both sLDA and \textsc{Scholar} treat the class label of each document as a response variable and jointly model both documents and their responses. We can observe that the latter is more effective in incorporating supervised information since both sLDA and \textsc{Scholar} outperform JST in general. But DIATOM gives significantly better results all over the baselines with the improvement over the best baseline model, \textsc{Scholar}, by 3-8\%. In our models, features used for sentiment classification are opinion-bearing topics. This shows that separating opinion topics from plot/neutral topics is beneficial for sentiment classification. We also observe that the contribution of the plot network to sentiment classification is dataset-dependent. The use of plot network largely boosts the sentiment classification accuracy by over 8\% on the Amazon dataset. But its effect is negligible on the other two datasets. 

When compared with traditional sentiment classification models such as SVM, we found that DIATOM outperforms SVM trained with various features on both IMDB and Amazon. But it performs slightly worse than SVM trained with TFIDF features with or without an additional incorporation of sentiment lexicon features. Nevertheless, DIATOM gives superior performance compared to SVM trained on LDA topic features in the range of 5-11\%, showing the effectiveness of using opinion topics for sentiment classification.

\begin{table*}[tb]\small
\centering
%\resizebox{\textwidth}{!}{ 
\scalebox{0.99}{
    \renewcommand{\arraystretch}{0.005}
    \begin{tabular}{@{}l@{}}
    \toprule 
    \multicolumn{1}{c}{\textbf{Plot/Neutral Topics}}  \\
    \midrule 
    
    \begin{tabularx}{\textwidth}{X}
    \textbf{Amazon - Topic 1} \emph{Dent, Gotham, City, Gordon, Bruce, Wayne, Harvey, Joker, Criminal, Nolan }  \\   \vspace{1pt}
    \rule{450pt}{0.5pt}
    \begin{enumerate}[nolistsep]
    \item Being imprisoned Batman has enough time to paint a gigantic flaming bat on a bridge while people are literally being executed on the hour.
    \item Batman gets with Catwoman... after how hard she sold him out?
    \end{enumerate}  \vspace{-7pt}
    \rule{450pt}{0.5pt}
    \textbf{Amazon - Topic 2} \emph{Gandalf, Frodo, Jackson, Tolkien, Dwarf, Fellowship, Peter, Orc, Ring, Hobbit}  \\   \vspace{1pt}
    \rule{450pt}{0.5pt}
    \begin{enumerate}[nolistsep]
    \item {[...]} the myriad inhabitants of Middle-earth, the legendary Rings of Power, and the fellowship of hobbits, elves, dwarfs, and humans--led by the wizard Gandalf (Ian McKellen) and the brave hobbit Frodo.
    \item This is the beginning of a trilogy; soon to be finalized.  \vspace{-7pt}
    \end{enumerate}\\
    %\rule{450pt}{0.5pt}
 \end{tabularx} \\
 
    \midrule
     \multicolumn{1}{c}{\textbf{Opinion-Bearing Topics}}\\
    \midrule 
   \begin{tabularx}{\textwidth}{X}
    \textbf{Amazon - Topic 1} \emph{Expectation, Quality, Definitely, Great, Good, Worth, Graphic, Predictable, Compare, Decent}\\   \vspace{1pt}
     \rule{450pt}{0.5pt}
    \begin{enumerate}[nolistsep]
     \item Action is good.
     \item Rachel Weisz was “mostly" good.
    \end{enumerate}   \vspace{-7pt}
    \rule{450pt}{0.5pt}
    \textbf{Amazon - Topic 2} \emph{Price, Shame, Service , Normally, Purchase, Connection, Greed, Stream, Watch, Frustrate}\\
     \rule{450pt}{0.5pt}
    \begin{enumerate}[nolistsep]
     \item This experience leaves me skeptical of the Amazon Prime video service.
     \item Look closely before purchasing.  \vspace{-8pt}
    \end{enumerate}    
    %\rule{450pt}{0.5pt}
    \end{tabularx} \\
 \bottomrule
 \end{tabular}
}
\caption{Example topics extracted by DIATOM from Amazon reviews and their associated most similar sentences.}
\label{tb:topic_examples}
\vspace{-1pt}
\end{table*}

%%%%%%%%%%%%%%%%%%%%%%%%
% Topic Disentanglement
%%%%%%%%%%%%%%%%%%%%%%%% 
\subsection{Topic Disentanglement}
\label{subs:topic_disent}

None of the aforementioned measures can, however, capture how opinion and plot topics are distributed. To this aim, we use topic labeling to assign a proxy label (\texttt{Positive}, \texttt{Negative}, \texttt{Plot}, \texttt{None}) to each topic and then measure the topic-disentanglement rate $\rho$ 
% (Eq. \ref{eq:disent_rate_def}) 
as the proportion of opinion-bearing topics with respect to the overall set of topics, complementary to the proportion of plot/neutral topics: $\rho = \frac{S}{S+K}$, with $S$ being the number of opinion topics and $K$ the number of plot/neutral topics.

% \begin{equation}
% \rho = \frac{S}{S+K} 
% \label{eq:disent_rate_def}
% \end{equation}

% We first compute the topic polarity with regard to the single topic $k_i$ and then, the rate of the overall set of topics $\mathcal{T}$.
For each topic, we first calculate its embedding by taking the normalized weighted average of the vectors of its top $N$ words:
$\vv{t_{z}} = \frac{1}{N}\sum_{i=1}^N \alpha_i\times\vv{w_i}$,
where $\alpha_i$ is the normalized distribution of word $w_i$ in topic $z$. %weights in the specific topic-word distribution.
We then retrieve the top 10 most similar sentences from the human-annotated sentence set measured by the cosine similarity between the topic embedding and each sentence embedding. The sentence embedding %$\vv{s_j}$ 
is computed using the Sentence-BERT encoder \cite{reimers2019}. %, an off-the-shelf encoder which has been proven to generate meaningful cost-effective representations for sentences. 
The most frequent label among the retrieved sentences is adopted as the topic's label.

\begin{figure}[!t]
\centering
\includegraphics[width=0.99\columnwidth]{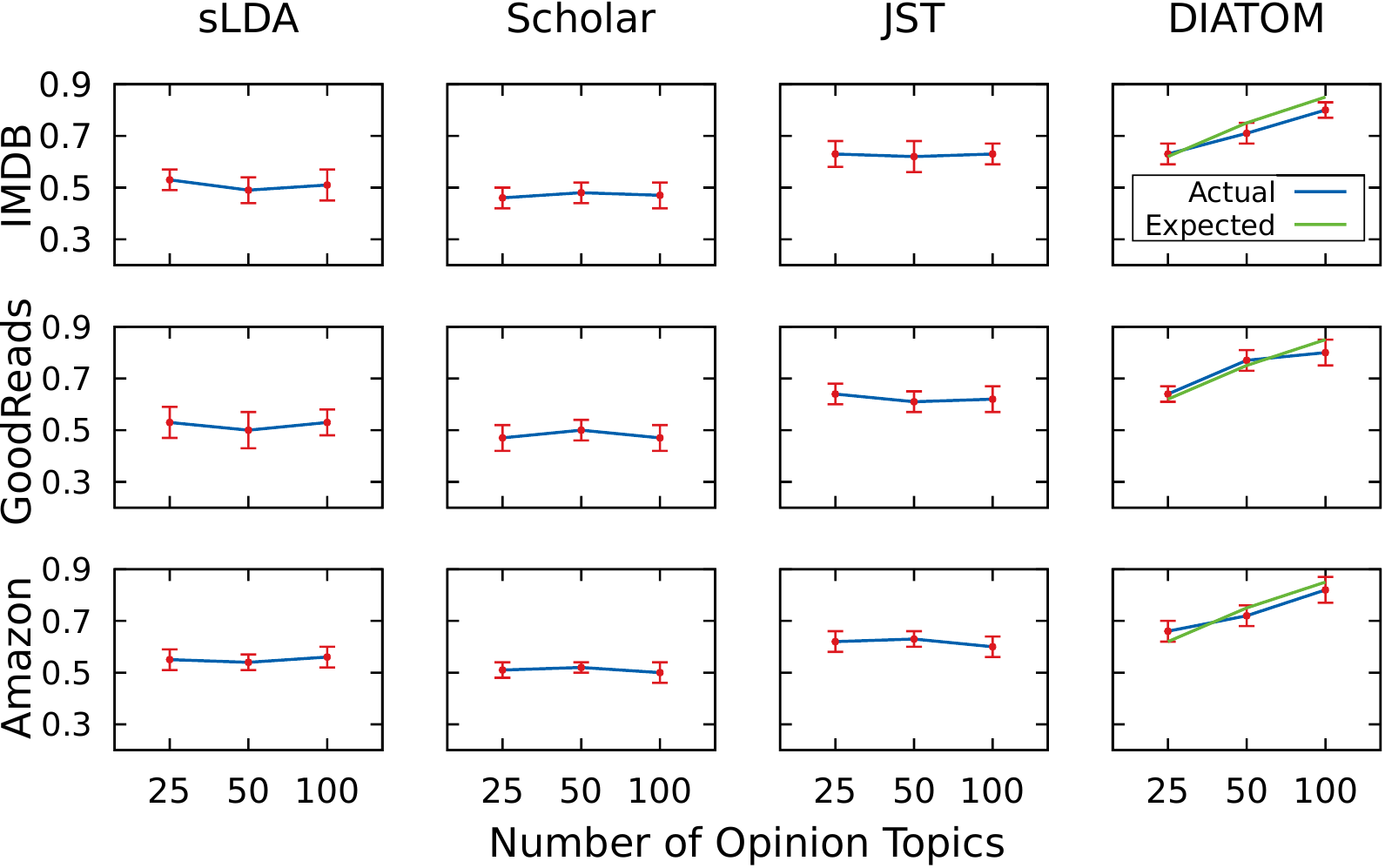}
\caption{Disentangling rate (\%) of topic models across different number of topics.} 
\label{fig:graph_disent_rate}
\end{figure}

To highlight the disentanglement capability of \textsc{DIATOM}, in Figure \ref{fig:graph_disent_rate}, we analyse how the proportion of opinion-bearing topics varies across standard and sentiment topic models. We notice that despite the signal from the document labels, sLDA and \textsc{Scholar} tend to produce topics rather balanced in terms of neutral and opinion-bearing topics. JST has a more skewed distribution towards opinion topics. DIATOM instead generates an actual proportion of opinion topics approaching the expected proportion set up by the model, demonstrating the capability to control the generation of plot and opinion-bearing topics.

In Table \ref{tb:topic_examples}, we show a set of topics grouped according to the disentanglement induced by DIATOM from Amazon\footnote{Topic results from IMDB and GoodRead can be found in the Appendix.}. For each topic, we report an excerpt of the most similar sentences retrieved. Aside from being overall coherent, we can guess %rather paradigmatic themes as the Topic 1 about %peace and war between countries, or 
more peculiar plots related for instance to `\emph{The Hobbit}' or `\emph{Batman}' as in the Plot/Neutral topics. 
%It is worth having a closer look at the IMDB-Topic 2, which despite the \emph{negative} theme of depression and suicide, the model is able to correctly gather those words under the same plot topic.
The opinion-bearing topics report a collection of commonly appreciated or critic aspects; some of them %are mainly collections of related adjectives with the same polarity (e.g. IMDB-Topic 1). %, while others 
are made up of mixed terms describing the issues and the associated experience (e.g. Topic 2). %Also, as already discussed in other works in literature,  sentences with negative aspects tend to be longer and more descriptive, compare to the ones with positive utterances. 

%%%%%%%%%%%%%%%%%%%%%%%%%%%
% t-SNE
\subsection{Visualization}
Another way to look at the disentangled topics is through the visualization of topic vectors. 

\begin{figure*}[!ht]
\centering
\captionsetup{width=1.0\linewidth}
\includegraphics[width=1.0\textwidth]{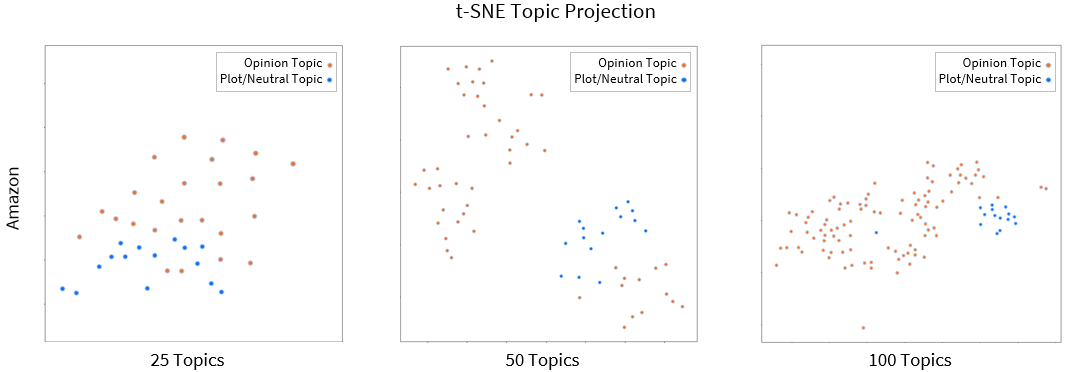}
\caption{Example of t-SNE projection for the Amazon dataset of the topic distribution for different number of topics. Color are assigned according to plot/neutral and opinion topics.} 
% \vspace{-12pt}
\label{fig:tsne_visualization}
\end{figure*}

As an example, we plot in Figure~\ref{fig:tsne_visualization} the 2-dimensional representation of the topic distributions projected by t-SNE for the Amazon dataset. 
Different colors represent different types of topics generated by DIATOM, namely plot/neutral in blue and opinion in red. We notice how consistently plot/neutral topics tend to cluster together across different number of topics, with the boundary close to polarized topics likely to share common features, as shown in Figure~\ref{fig:ex_diatom} in which the plot topic and the negative topic share a common word `\emph{Dwarf}'.

%\clearpage

%%%%%%%%%%%%%%%%%%%%%%%%%%%
%\subsection{Further Discussion}
\subsection{Ablation Study} 

We report in Table \ref{tb:ablation_results} the results of the ablation study on DIATOM. We observe that removing the orthogonal regularization has a limited effect on sentiment classification, but causes a fluctuation on topic coherence and a clear drop in topic uniqueness. % thus corroborating the idea that such a regularizer drives the model towards higher topic diversity regardless of the topic coherence. 
A significant classification performance drop is observed by removing the sentiment classifier, which essentially reduces DIATOM to an unsupervised model.
Removing both the orthogonal regularization and the sentiment classifier %decreases simultaneously the generalisation capability and the supervised signal, with 
shows a major negative impact on both accuracy and the topic's quality. 
Finally, we assess the influence of the plot network (\textsection \ref{s:diatom_architecture}), and while we do not notice any consistent impact across the datasets in terms of sentiment classification, the 
% quality of 
topics has a notable drop in terms of coherence and diversity.

\begin{table}[!h]
\centering
% \setlength{\tabcolsep}{15pt}
% \resizebox{\textwidth}{!}{ 
\resizebox{\columnwidth}{!}{
\begin{tabular}{@{}l@{}lccc@{}}
\toprule
\textbf{Datasets} &  \textbf{Models} & \textbf{Accuracy}  & \textbf{TC $\,/\,$ TU}\\ 
    \midrule
    IMDB  & \textsc{DIATOM}     & $0.726$ \text{\footnotesize $\pm\,0.03$} & 0.639 $/$ 38.1  \\ 
      &  -- w/o orth reg.       & $0.723$ \text{\footnotesize $\pm\,0.01$} & 0.582 $/$ 27.5 \\
      &  -- w/o sent. class.    & $0.491$ \text{\footnotesize $\pm\,0.03$} & 0.601 $/$ 35.4 \\
      &  -- w/o both            & $0.478$ \text{\footnotesize $\pm\,0.03$} & 0.544 $/$ 25.4 \\
      &  -- w/o Plot Net         & $0.734$ \text{\footnotesize $\pm\,0.03$} & 0.603 $/$ 36.7 \\
    %   \midrule
      \midrule
    GoodReads $\, $ & \textsc{DIATOM} & $0.704$  \text{\footnotesize $\pm\,0.02$} & 0.634 $/$ 52.9  \\
      &  -- w/o orth. reg.      & $0.681$ \text{\footnotesize $\pm\,0.02$} & 0.612 $/$ 41.1 \\
      &  -- w/o sent. class.    & $0.446$ \text{\footnotesize $\pm\,0.02$} & 0.638 $/$ 47.6 \\
      &  -- w/o both            & $0.410$ \text{\footnotesize $\pm\,0.02$} & 0.552 $/$ 39.6 \\
      &  -- w/o Plot Net          & $0.695$ \text{\footnotesize $\pm\,0.03$} & 0.615 $/$ 49.3 \\
    %   \midrule
      \midrule
    Amazon  &  \textsc{DIATOM}  & $0.686$ \text{\footnotesize $\pm\,0.02$} & 0.598 $/$ 82.3 \\
      &  -- w/o orth reg.       & $0.682$ \text{\footnotesize $\pm\,0.01$} & 0.605 $/$ 55.3 \\
      &  -- w/o sent. class.    & $0.601$ \text{\footnotesize $\pm\,0.03$} & 0.573 $/$ 76.9 \\
      &  -- w/o both            & $0.548$ \text{\footnotesize $\pm\,0.03$} & 0.567 $/$ 52.1 \\
      &  -- w/o Plot Net          & $0.603$ \text{\footnotesize $\pm\,0.02$} & 0.584 $/$ 78.3 \\
    \bottomrule
  \end{tabular}
%   }
}
\caption{Ablation study over DIATOM by removing the orthogonal regularization, the sentiment classifier or just the auxiliary Plot Network.}
\label{tb:ablation_results}
\end{table}

%%%%%%%%%%%%%%%%%%%%%%%%%%%
\subsection{Further Discussion}
%\noindent\textbf{Limitation and extension}. 
Although the adversarial mechanism implemented in DIATOM is rather effective in disentangling opinion and neutral/plot topics, at times the opinion topics could exhibit terms of mixed polarities. An additional adversarial mechanism can be a viable solution at the cost of increasing the model's complexity.

%Moreover,
In our current model, the latent plot topics $\bm{z}_a$ extracted from reviews are encouraged to have a similar discriminative power as the latent topic $\bm{z}_d$ learned from plots directly for predicting the plots. It is also possible to impose a Gaussian prior centred on $\bm{z}_d$ for the latent plot topics in reviews instead of using the Gaussian prior of zero mean and unit variance. 

% Another approach is to replace the plot classifier with a discriminator as typically used in GAN training that the learned plot topics from different sources (reviews and plots) are competing with each other to confuse the discriminator. 

While we focus on separating opinion topics from plot or neutral ones in movie and book reviews in this work, our proposed framework can be applicable in other scenarios. For example, in veracity classification of Twitter rumours, we want to disentangle latent factors which are indicative of veracity of tweets from those which are event-related. Our proposed framework provides a potential solution to it.

\section{Conclusions}
\label{s:conclusion}
We have described DIATOM, a new neural topic model to generate disentangled topics through the combination of VAE and adversarial learning. The results on the novel \textsc{MOBO} dataset %highlight the benefit of such an approach leading to topics with higher interpretability 
show that DIATOM generates better topics in terms of both topic coherence and topic uniqueness, and can  
%We further discussed the model capability to consistently 
disentangle opinion-bearing topics from plot/neutral ones. % measuring the introduced disentangling rate. 
Finally, we have identified some existing limitations and provided viable solutions to be explored in the future.

\section*{Acknowledgements}
Our gratitude goes to all the annotators and to Fabio Miceli and Gianluca Pergola for their relentless effort.
This work is funded by the EPSRC (grant no. EP/T017112/1, EP/V048597/1). YH is supported by a Turing AI Fellowship funded by the UK Research and Innovation (UKRI) (grant no. EP/V020579/1).

\bibliography{anthology,custom}
\bibliographystyle{acl_natbib}

%%%%%%%%%%%%%%%%%%%%%%%%%%%%
%  APPENDIX
%%%%%%%%%%%%%%%%%%%%%%%%%%%%
%\newpage
\clearpage

%\onecolumn
\appendix

\setcounter{table}{0}
\makeatletter 
\renewcommand{\thetable}{A\@arabic\c@table}
\makeatother

\balance
\section{Appendix}
\label{sec:appendix}

%%%%%%%%%%%%%%%%%%%%%%%%%% 
\subsection{Training Details}

\paragraph{Hyperparameters} We tune the models' hyperparameters on the development set via a random search over combinations of learning rate $\lambda \in [0.001, 0.5]$, dropout $\delta \in [0.0,0.6]$ and topic vector size $\gamma_t \in [25,50,100,200]$.
%\footnote{\gab{One may argue that the number of topics across models should be set to be overall exactly the same. However, for a fixed number of topics there is an extremely high number of possible combinations of polarised and neutral topics; preliminary results showed that results are less fluctuating when fixing the number of neutral topics.}}.
% 
Encoder and decoder are configured following \cite{Srivastava17}. The hidden representation of documents is set to $100$ and sentiment classifier's hidden size to 50. Matrices are randomly initialized with the Xavier and sparse methods \cite{Glorot10, Martens10}.
We employ the Adam optimizer \cite{Kingma2014}, set the batch size to 64 and apply batch normalization as additional regularizer \cite{Cooijmans17}.

% TODO: Add symbols about the model components
\paragraph{Sequential Unfreezing} Instead of simultaneously training all the model components, we unfreeze
them sequentially. We first freeze the sentiment classifier and update only the autoencoder. At the $e_{th}$ epoch, we unfreeze the sentiment classifier uniquely on the polarized features to let the classifier training. Finally, at the $(e+n)_{th}$ epoch, we unfreeze the adversarial mechanism to drive the generation of neutral features fooling the classifier. 
We follow an analogous approach with regards to the plot classifier.
The values of $e$ and $a$ are treated as hyperparameters and chosen through the random search. We found that the sequential unfreezing scheme leads to better topic disentanglement.

\subsection{Example Topic Results}

In Table \ref{tb:topic_examples_full}, we show a set of topics grouped according to the disentanglement induced by DIATOM from IMDB and GoodReads. For each topic, we report an excerpt of the most similar sentences retrieved. Aside from being overall coherent, we can guess rather paradigmatic themes as the Topic 1 about peace and war between countries in IMDB-Topic 1. %, or more peculiar plots related for instance to `\emph{The Hobbit}' or `\emph{Batman}' as in the Plot/Neutral topics. 
It is worth having a closer look at the IMDB-Topic 2, which despite the \emph{negative} theme of depression and suicide, the model is able to correctly gather those words under the same plot topic. The opinion-bearing topics report a collection of commonly appreciated or critic aspects; some of them are mainly collections of related adjectives with the same polarity (e.g. IMDB-Topic 1). %, while othe

\begin{table*}[htbp]\small
\centering
%\resizebox{\textwidth}{!}{ 
\scalebox{0.99}{
    \renewcommand{\arraystretch}{0.005}
    \begin{tabular}{@{}l@{}}
    \toprule %\midrule
    \multicolumn{1}{c}{\textbf{Plot/Neutral Topics}}  \\
    \midrule %\midrule
    
    \begin{tabularx}{\textwidth}{X}
    \textbf{IMDB - Topic 1} \emph{Government, Country, Peace, Information, Free, Plane, Theory, Anti, Soldier, Hitler}\\
     \midrule
    \begin{enumerate}[nolistsep]
    \item Groundbreaking in the realm of socially relevant drama, it dealt with issues such as abortion, domestic violence, student protest, child neglect, illiteracy, slumlords, the anti-war movement, [...].
    \item This effort by Charlie ultimately evolves into a major portion of the U.S. foreign policy known as the Reagan Doctrine, under which the U.S. expanded assistance beyond just the [...].
    \end{enumerate}  \vspace{-7pt}
    \rule{450pt}{0.5pt}
    \textbf{IMDB - Topic 2} \emph{Window, Hospital, Apartment, Suicide, Commit, Pitt, Serial, Strange, Killer, Mental}  \\   \vspace{1pt}
    \rule{450pt}{0.5pt}
    \begin{enumerate}[nolistsep]
    \item Even re-think why two boys/young men would do what they did - commit mutual suicide via slaughtering their classmates.
    \item It's the patented scene where the killer creeps up behind the victim.
    \end{enumerate}  \vspace{-7pt}
    \rule{450pt}{0.5pt}
    \textbf{GoodReads - Topic 1} \emph{Cure, Plague, Trial, Betray, Thomas, Secret, Dashner, Ball, Betrayal, Wicked}  \\   \vspace{1pt}
    \rule{450pt}{0.5pt}
    \begin{enumerate}[nolistsep]
    \item Blaming Cinder for her daughter's illness, Cinder's stepmother volunteers her body for plague research, an "honor" that no one has survived.
    \item By age thirteen, she has undergone countless surgeries, transfusions, and shots so that her older sister, Kate, can somehow fight the leukemia that has plagued her since childhood.
    \end{enumerate}  \vspace{-7pt}
    \rule{450pt}{0.5pt}
    \textbf{GoodReads - Topic 2} \emph{Teenager, Fault, Illness, Mental, Depression, Maddy, Grief, Bully, Topic, Greg}  \\   \vspace{1pt}
    \rule{450pt}{0.5pt}
    \begin{enumerate}[nolistsep]
    \item She's got a lot of mental strength, having been ostracized for most of her life.
    \item  She went through a divorce, a crushing depression, another failed love, and the eradication of everything she ever thought she was supposed to be.
    \end{enumerate} \vspace{-7pt}
%    \rule{450pt}{0.5pt}
%    \textbf{Amazon - Topic 1} \emph{Dent, Gotham, City, Gordon, Bruce, Wayne, Harvey, Joker, Criminal, Nolan }  \\   \vspace{1pt}
%    \rule{450pt}{0.5pt}
%    \begin{enumerate}[nolistsep]
%    \item Being imprisoned Batman has enough time to paint a gigantic flaming bat on a bridge while people are literally being executed on the hour.
%    \item Batman gets with Catwoman... after how hard she sold him out?
%    \end{enumerate}  \vspace{-7pt}
%    \rule{450pt}{0.5pt}
%    \textbf{Amazon - Topic 2} \emph{Gandalf, Frodo, Jackson, Tolkien, Dwarf, Fellowship, Peter, Orc, Ring, Hobbit}  \\   \vspace{1pt}
%    \rule{450pt}{0.5pt}
%    \begin{enumerate}[nolistsep]
%    \item {[...]} the myriad inhabitants of Middle-earth, the legendary Rings of Power, and the fellowship of hobbits, elves, dwarfs, and humans--led by the wizard Gandalf (Ian McKellen) and the brave hobbit Frodo.
%    \item This is the beginning of a trilogy; soon to be finalized.  \vspace{-7pt}
%    \end{enumerate}\\
    \end{tabularx}\\
    
    \midrule%\midrule
     \multicolumn{1}{c}{\textbf{Opinion-Bearing Topics}}\\
    \midrule %\midrule
    \begin{tabularx}{\textwidth}{X}
    \textbf{IMDB - Topic 1} \emph{Badly, Stock, Remove, Poorly, Hype, Ridiculous, Insult, Disaster, Excuse, Lame}\\   \vspace{1pt}
     \rule{450pt}{0.5pt}
    \begin{enumerate}[nolistsep]
    \item I can't imagine how anyone could have read this badly written script and given it the greenlight.
    \item Although there has obviously been a lot of money spent on them the numbers are badly staged and poorly photographed.
    \end{enumerate}   \vspace{-7pt}
    \rule{450pt}{0.5pt}
    \textbf{IMDB - Topic 2} \emph{Exceptional, Recommend, Excellent, Craft, Believable , Overlook, Vhs, Solid, Festival, Amaze }\\   \vspace{1pt}
     \rule{450pt}{0.5pt}
    \begin{enumerate}[nolistsep]
     \item Overall, this is a good film and an excellent adaption.
     \item It's great acting, superb cinematography and excellent writing.
    \end{enumerate}  \vspace{-7pt}
    \rule{450pt}{0.5pt}
    \textbf{GoodReads - Topic 1} \emph{Negative, Judge, Note, Pretend, Embarrass, Quality, Extreme, Guilty, Fake, Borrow}\\   \vspace{1pt}
     \rule{450pt}{0.5pt}
    \begin{enumerate}[nolistsep]
     \item Can you give something negative stars?
     \item And while it must be hard reading negative reviews you need to be able to deal with this in a graceful way (no one likes a sore loser).
    \end{enumerate}  \vspace{-7pt}
    \rule{450pt}{0.5pt}
    \textbf{GoodReads - Topic 2} \emph{Teen, Nice, Normally, Little, Genre, Amuse, Theme, Enjoyment, Blow, Reread}\\   \vspace{1pt}
     \rule{450pt}{0.5pt}
    \begin{enumerate}[nolistsep]
     \item What would have made the book a lot more fun to read was more meatier characters in the other girls.
     \item But I feel like that was part of the fun of it.
    \end{enumerate}   \vspace{-7pt}
%    \rule{450pt}{0.5pt}
%    \textbf{Amazon - Topic 1} \emph{Expectation, Quality, Definitely, Great, Good, Worth, Graphic, Predictable, Compare, Decent}\\   \vspace{1pt}
%     \rule{450pt}{0.5pt}
%    \begin{enumerate}[nolistsep]
%     \item Action is good.
%     \item Rachel Weisz was “mostly" good.
%    \end{enumerate}   \vspace{-7pt}
%    \rule{450pt}{0.5pt}
%    \textbf{Amazon - Topic 2} \emph{Price, Shame, Service , Normally, Purchase, Connection, Greed, Stream, Watch, Frustrate}\\
%     \rule{450pt}{0.5pt}
%    \begin{enumerate}[nolistsep]
%     \item This experience leaves me skeptical of the Amazon Prime video service.
%     \item Look closely before purchasing.  \vspace{-8pt}
%    \end{enumerate}
    \end{tabularx} \\
 \bottomrule
 \end{tabular}
}
\caption{Example topics extracted by DIATOM from IMDB and GoodReads and their associated most similar sentences.}
\label{tb:topic_examples_full}
\end{table*}

\end{document}